\documentclass[runningheads]{llncs}

\usepackage{eccv}

\usepackage{eccvabbrv}

\usepackage{graphicx}
\usepackage{booktabs}
\usepackage{algorithm}
\usepackage{algorithmic}
\usepackage{multirow}
\usepackage{threeparttable}
\usepackage{amssymb}
\usepackage{xcolor}
\usepackage{pifont}
\newcommand{\cmark}{\ding{51}}%
\newcommand{\xmark}{\ding{55}}%
\usepackage{amsmath}
\usepackage{adjustbox}
\definecolor{citecolor}{HTML}{0071bc}
\definecolor{ourscolor}{HTML}{c2d1e5}
\usepackage{colortbl}  
\usepackage[misc]{ifsym}
\definecolor{darkcandyapplered}{rgb}{0.64, 0.0, 0.0}
\definecolor{burgundy}{rgb}{0.5, 0.0, 0.13}
\definecolor{carnelian}{rgb}{0.7, 0.11, 0.11}

\definecolor{tabhighlight}{HTML}{ffffff} 
\definecolor{tabhighlightric}{HTML}{ffffff} 

\usepackage{amsmath,amsfonts,bm}

\def\eqref#1{equation~\ref{#1}}

\def\1{\bm{1}}

\DeclareMathAlphabet{\mathsfit}{\encodingdefault}{\sfdefault}{m}{sl}
\SetMathAlphabet{\mathsfit}{bold}{\encodingdefault}{\sfdefault}{bx}{n}

\makeatletter
\def\blfootnote{\gdef\@thefnmark{}\@footnotetext}
\makeatother

\usepackage[accsupp]{axessibility}  

\usepackage{hyperref}

\usepackage{orcidlink}

\begin{document}

\title{When Pedestrian Detection Meets Multi-Modal Learning: Generalist Model and Benchmark Dataset}

\titlerunning{MMPedestron}

\author{
Yi Zhang\inst{2}\orcidlink{0009-0006-7303-7594} \and 
Wang Zeng\inst{2}\orcidlink{0000-0003-1562-6332} \and
Sheng Jin\inst{3,2}\orcidlink{0000-0001-5736-7434} \and
Chen Qian\inst{1,2}\orcidlink{0000-0002-8761-5563} \textsuperscript{\Letter} \\
Ping Luo\inst{3,4}\orcidlink{0000-0002-6685-7950} \and
Wentao Liu\inst{2}\orcidlink{0000-0001-6587-9878} 
}

\authorrunning{Y. Zhang et al.}

\institute{$^{1}$ Tsinghua University \quad
$^{2}$ SenseTime Research and Tetras.AI  \\
$^{3}$ The University of Hong Kong \quad
$^{4}$ Shanghai AI Laboratory \\
\email{yizhang.bham.uk@outlook.com, qianc18@mails.tsinghua.edu.cn}}
\maketitle

\begin{abstract}

Recent years have witnessed increasing research attention towards pedestrian detection by taking the advantages of different sensor modalities (\eg RGB, IR, Depth, LiDAR and Event). However, designing a unified generalist model that can effectively process diverse sensor modalities remains a challenge. This paper introduces MMPedestron, a novel generalist model for multimodal perception. Unlike previous specialist models that only process one or a pair of specific modality inputs, MMPedestron is able to process multiple modal inputs and their dynamic combinations. The proposed approach comprises a unified encoder for modal representation and fusion and a general head for pedestrian detection. We introduce two extra learnable tokens, \ie MAA and MAF, for adaptive multi-modal feature fusion. In addition, we construct the MMPD dataset, the first large-scale benchmark for multi-modal pedestrian detection. This benchmark incorporates existing public datasets and a newly collected dataset called EventPed, covering a wide range of sensor modalities including RGB, IR, Depth, LiDAR, and Event data. With multi-modal joint training, our model achieves state-of-the-art performance on a wide range of pedestrian detection benchmarks, surpassing leading models tailored for specific sensor modality. For example, it achieves 71.1 AP on COCO-Persons and 72.6 AP on LLVIP. Notably, our model achieves comparable performance to the InternImage-H model on CrowdHuman with $30\times$ smaller parameters. Codes and data are available at \url{https://github.com/BubblyYi/MMPedestron}.
\blfootnote{\Letter~: Corresponding author.}

\end{abstract}
    
\section{Introduction}
\label{sec:intro}

\begin{figure}[!htb]
    \centering
    \begin{minipage}{0.43\textwidth}
    \centering
    \includegraphics[width=0.84\textwidth]{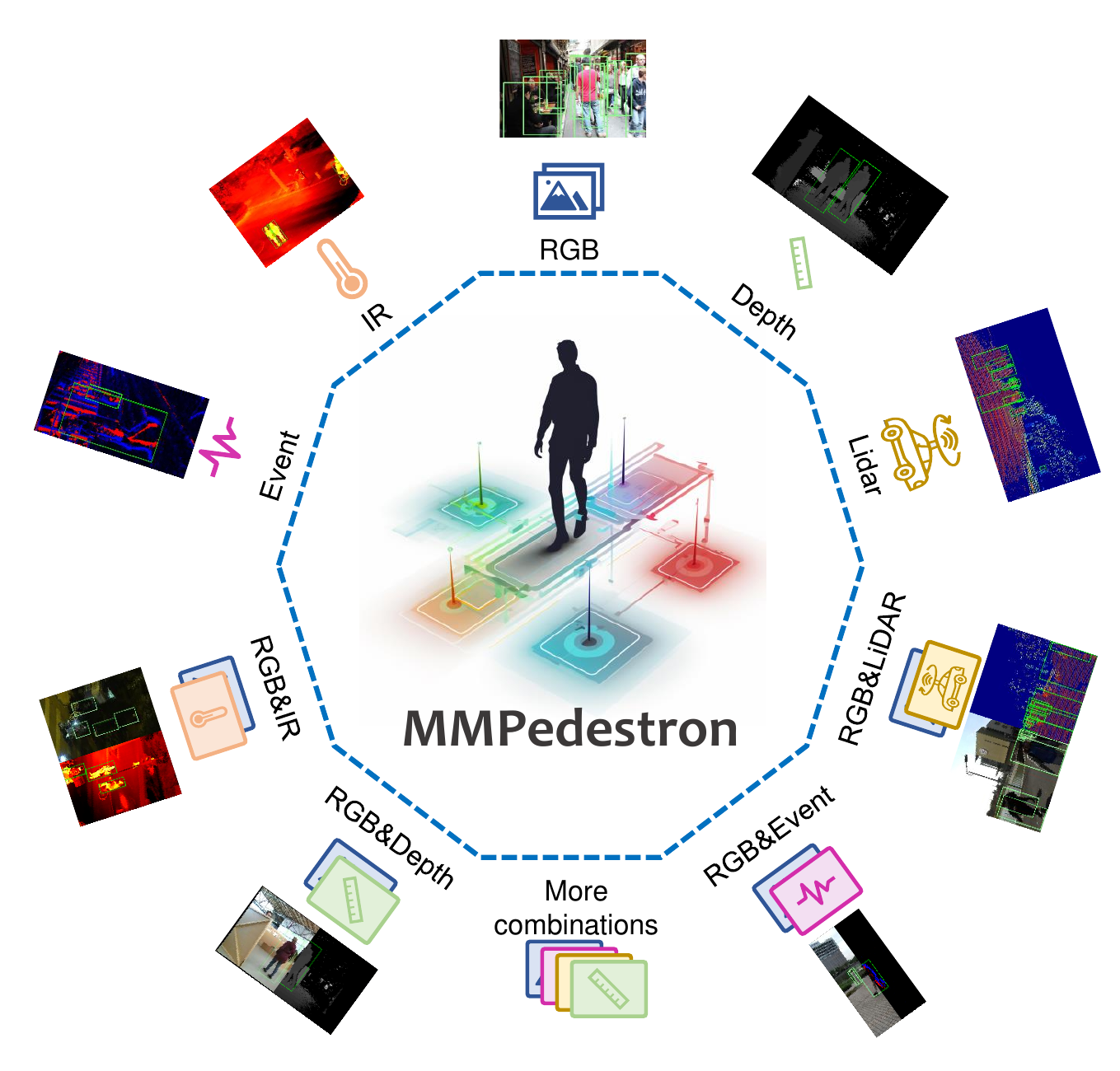}
    \caption{MMPedestron unifies diverse modality inputs, including RGB, IR, Event, Depth and LiDAR, for pedestrian detection.} 
  \label{fig:intro}
    \end{minipage}%
    \hspace{4mm}
    \begin{minipage}{0.46\textwidth}
        \centering
    \includegraphics[width=0.99\textwidth]{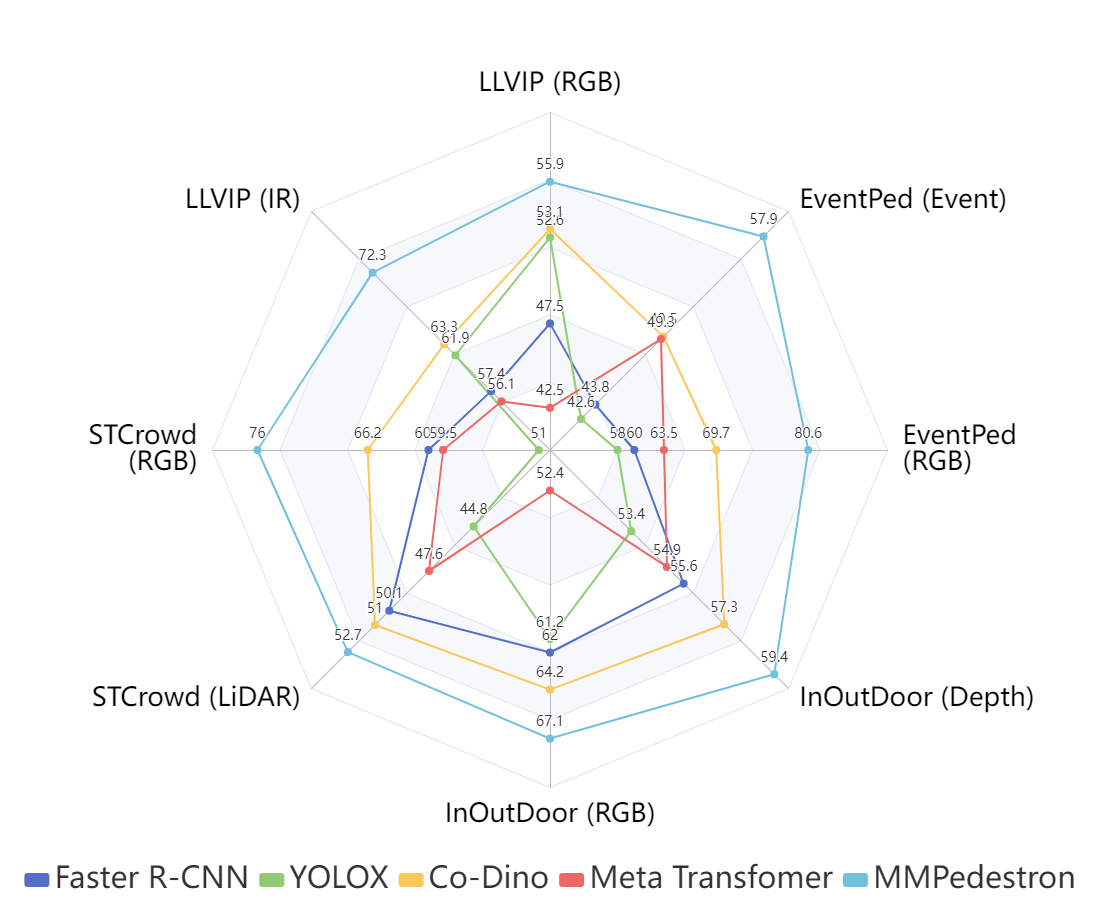}
    \caption{Performance on diverse datasets and modalities. MMPedestron outperforms leading models trained on the specific dataset and modality.}
    \label{fig:unimodal}
    \end{minipage}
    \vspace{-3mm}
\end{figure}

Pedestrian detection~\cite{dollar2011pedestrian} has long been a hot research topic in computer vision due to its various applications, including autonomous driving, robotics and video surveillance. 
Traditional pedestrian detection mostly focus on single-modal RGB images as the input. However, RGB based pedestrian detection methods face great challenges in complex scenarios (\eg background clutter and adverse lighting conditions).
With the rapid development and application of sensing hardware, multi-modal learning has attracted increasing research attention. 
Different types of sensors can supply RGB images with rich complementary information and bring remarkable benefits for pedestrian detection. 
For example, Infrared Radiation (IR) sensor detects heat radiation of pedestrians which is helpful for detection in a dark environment. 
Time of Flight (ToF) and LiDAR sensors provide additional depth information of the scene. 

Previous studies mainly focus on designing specific models for one single or a pair of modality inputs. The development of a unified model that can effectively incorporate various sensor modalities in multi-modal pedestrian detection poses several challenges.
Firstly, existing benchmarks for pedestrian detection primarily focus on a single or a pair of sensor modalities, lacking a comprehensive benchmark that can fairly and comprehensively evaluate various methods across diverse application scenarios.
Secondly, previous multi-modal fusion methods are often tailored for specific modality pairs (\eg RGB-D or RGB-T), yet hard to be extended to operate with other modality combinations. 
For example, those models trained for RGB-D data are not applicable to the RGB-T data. 
Consequently, multiple models are required to deal with different modality combinations, resulting in unnecessary system complexity and inefficiency.
In addition, previous fusion methods assume the availability of all modalities and do not account for scenarios where certain modalities may be missing, exacerbating the problem. 
Lastly, different modality-specific pedestrian datasets are collected from various domains and designed for specific application scenarios (\eg LLVIP~\cite{jia2021llvip} for surveillance viewpoints, Waymo~\cite{sun2020scalability} for automobiles, InOutDoor~\cite{mees2016choosing} for robotics). As a result, previous pedestrian detectors trained on one specific modality lack generalization capabilities across different domains.

In this paper, we make contributions to the field of multi-modal pedestrian detection by introducing both a benchmark dataset and a generalist model.
Firstly, we address the lack of a comprehensive benchmark for multi-modal pedestrian detection by constructing the MMPD benchmark. This benchmark dataset is derived from existing public datasets~\cite{lin2014microsoft,shao2018crowdhuman,jia2021llvip,mees2016choosing,cong2022stcrowd}. 
To address the lack of paired RGB-Event data in the community and to enhance the MMPD benchmark's diversity, we also introduce a new RGB-Event pedestrian detection dataset called EventPed.
Our MMPD dataset is diverse in two aspects. \textbf{(1) Modality.} MMPD dataset covers multiple sensor modalities, such as RGB, IR, Depth, LiDAR, and Event data, and diverse modality combinations, including RGB+IR, RGB+Depth, RGB+LiDAR, and RGB+Event.
\textbf{(2) Scenario.} Unlike previous datasets collected under a specific scenario, MMPD encompasses various scenarios, including surveillance, automobile, robotics, outdoor and indoor environments.
The diversity of modality and scenario make it possible to develop and evaluate the generalist multi-modal pedestrian detection model.

We propose MMPedestron, a generalist multi-modal pedestrian detection model designed to handle diverse input modalities and scenarios.
MMPedestron consists of a unified multi-modal encoder and a detection head.
The unified encoder transfer multi-modal input to vision tokens, which are combined with a  Modality Aware Fuser (MAF) and a Modality Aware Abstractor (MAA) token to form a hybrid token sequence. The hybrid token sequence is processed by transformer blocks and transferred to unified tokens by modality unifier module.
The unified tokens are then passed to the detection head for final predictions.
By training on data with different modalities, MMPedestron achieves state-of-the-art performance on various pedestrian detection benchmarks, surpassing models tailored for specific sensor modalities. 
Furthermore, we highlight several noteworthy properties of MMPedestron:
(1) \textbf{Flexibility:} MMPedestron exhibits the capability to handle diverse input modalities and their dynamic combinations, allowing for versatility in different applications.
(2) \textbf{Scalability:} With efficient weight sharing, MMPedestron can seamlessly accommodate an increasing number of modalities without a proportional growth in parameters, demonstrating excellent scalability.
(3) \textbf{Generalization ability:} The diversity of the MMPD dataset enables MMPedestron to exhibit strong generalization abilities across various domains and scenarios.

Our main contributions can be summarized as follows:
\begin{itemize}

\item The introduction of MMPD dataset, a large-scale multi-modal pedestrian detection benchmark, which serves as a standardized evaluation platform for multi-modal pedestrian detection methods.

\item Pioneering the concept of generalist multi-modality pedestrian detection through the development of the MMPedestron model. This model is designed to handle diverse input modalities and scenarios, showcasing remarkable flexibility, scalability, and generalization ability.

\item Experimental results showcase that our model achieves state-of-the-art performance across a wide range of pedestrian detection benchmarks, outperforming current leading models tailored for specific sensor modality.

\end{itemize}

\section{Related Works}

\subsection{Multi-Modal Object Detection}
While RGB images provide substantial texture information and details for pedestrian detection, the inclusion of multi-modal data is desirable to achieve more reliable results in challenging conditions, such as extreme lighting, occlusions, and fast motion. 
Existing methods have developed various strategies to fuse information from multiple modalities, namely early-fusion, late-fusion, and mid-fusion.
\textbf{Early-fusion} (pixel fusion) concatenates data from different modalities and process it with regular object detectors~\cite{ren2015faster, lin2017feature, redmon2016you}. 
\textbf{Late-fusion} (decision fusion) feeds the inputs of two modalities separately into two unimodal object detection models, and then fusing predicted bounding boxes using statistical methods~\cite{chen2022multimodal, li2018multispectral, takumi2017multispectral}. 
Early-fusion and late-fusion approaches are straightforward, however they ignore the correlations between modalities. 
\textbf{Mid-fusion} (feature fusion) fuses the features extracted from multiple modalities and predicts bounding boxes from the fused feature. Most research on multi-modal object detection focuses on mid-fusion~\cite{zhang2021guided, liu2016multispectral, qingyun2022cross, cao2023multimodal, qingyun2021cross, tomy2022fusing}, as this strategy enables the deep exploration of the correlations between modalities.

Our MMPedestron model falls within the mid-fusion category. Previous mid-fusion methods usually use separated branches for different modalities. However, we use a unified encoder for all modalities, demonstrating better scalability.
In addition, previous fusion approaches primarily focus on bi-modal features and do not consider diverse modal combinations. In contrast, our model offers flexibility in handling diverse combinations of modalities and scenarios.

\subsection{Multi-Modal Benchmarks}
To facilitate the development of generalist multi-modality pedestrian detection models, it is crucial to have diverse data comprising various modalities. As depicted in Table~\ref{tab:dataset}, while there exists large repositories of annotated RGB-based datasets, there are much fewer annotated data of other modalities (\eg Depth), and even scarcer annotations of modality combinations (\eg RGB + Depth). 
More importantly, existing multi-modal benchmarks for pedestrian detection typically consist of only a single pair of modalities, \eg LLVIP~\cite{jia2021llvip} and InOutDoor~\cite{mees2016choosing}. In contrast, MMPD dataset which integrates multiple public datasets, encompasses five modalities and four distinct modal combinations. 

\begin{table}[tb]
\caption{Overview of representative multi-modal pedestrian detection datasets. ``\# Img'' means the number of total images. ``Label'' includes manual labels or pseudo-labels generated by models. MMPD has paired modality data covering all modalities.}
\begin{center}
\vspace{-5mm}
\scalebox{0.6}{
\setlength{\tabcolsep}{10pt}
\begin{tabular}{l|ccccccc}
    \hline
    Dataset & \# Img  & Label  & RGB & IR & LiDAR & Depth & Event  \\ \hline
    \textit{Caltech}~\cite{dollar2009pedestrian}  & 250K & Manual & \cmark   \\
    \textit{CityPersons}~\cite{zhang2017citypersons} & 5K & Manual &  \cmark \\
    \textit{CrowdHuman}~\cite{shao2018crowdhuman}  & 24K & Manual & \cmark  \\
    \textit{Objects365-Persons}~\cite{shao2019objects365}        & 133K & Manual &  \cmark \\ 
    \textit{COCO-Persons}~\cite{lin2014microsoft}  & 66K & Manual &  \cmark \\ \hline
    \textit{LLVIP}~\cite{jia2021llvip}       & 15K & Manual & \cmark  &   \cmark \\ 
    \textit{M3FD}~\cite{liu2022target}     & 4K & Manual & \cmark  &   \cmark \\ 
    \textit{FLIR}~\cite{FLIR_dataset}     & 26K & Manual &  \cmark  &   \cmark \\ 
    \hline 
    \textit{STCrowd}~\cite{cong2022stcrowd}  & 8K & Manual &   \cmark  &  &  \cmark  \\ 
    \textit{Waymo}~\cite{sun2020scalability}            & 36K & Manual & \cmark  &  &  \cmark  \\ 
    \hline 
    \textit{InOutDoor}~\cite{mees2016choosing} & 7K & Manual &  \cmark  & & & \cmark \\ 
    \textit{MobilityAids}~\cite{vasquez2017deep} & 17K & Manual &  \cmark  & & & \cmark \\ 
    \hline \textit{DSEC}~\cite{gehrig2021dsec,tomy2022fusing} & 11K  & Pseudo  & \cmark &  & &  & \cmark \\ 
    \textit{PEDRo}~\cite{bbpprsPedro2023} & 27K & Manual &  &  & &  & \cmark \\ 
    \textit{EventPed} (Ours) & 9K & Manual & \cmark  &  & &  & \cmark \\ 
    \hline 
    \textit{MMPD} (Ours) & 260K & Manual & \cmark  & \cmark & \cmark & \cmark & \cmark \\ \hline
\end{tabular}
}
\end{center}
\label{tab:dataset}
\vspace{-8mm}
\end{table}

\textbf{Event-based datasets for pedestrian detection.} 
The advantages in handling challenging lighting conditions, high motion, and low latency make event data well-suited for pedestrian detection~\cite{de2020large, perot2020learning, gehrig2021dsec}. However, due to the challenges in data collection and annotation, the availability of annotated event data is significantly limited compared to other common modalities.
GEN1 dataset~\cite{de2020large} and PEDRo dataset~\cite{bbpprsPedro2023} offer manual annotations, however they have relatively low image resolution and lack paired RGB images.
DSEC~\cite{gehrig2021dsec} offers paired RGB and event data without object bounding box annotations. A recent work~\cite{tomy2022fusing} proposes an automated labeling protocol to generate box annotations for DSEC, but the utilized data is not publicly released.
In contrast, our proposed EventPed dataset overcomes these limitations by providing high-resolution RGB-Event pairs collected in diverse environments, along with comprehensive manual annotations for pedestrian detection.

\subsection{Generalist Model}
\textbf{Multi-modal generalist model.}
The emergence of unified models that incorporate multiple modalities has garnered significant attention due to their exceptional performance across various tasks. 
ImageBind~\cite{girdhar2023imagebind} learns a joint embedding across six different modalities through data pairs of image and other modalities.
LanguageBind~\cite{zhu2023languagebind} further improves the joint embedding space by considering language as the bind modality.
While these works utilize separate encoders for different modalities, our MMPedestron employs a shared encoder for all modalities.  
Meta-Transformer~\cite{zhang2023meta} leverages a frozen encoder trained on RGB images to perform the perception of multiple modalities. However, Meta-Transformer processes only a single modality per task. In contrast, MMPedestron is capable of handling diverse combinations of multiple modalities. Moreover, Meta-Transformer is exclusively trained on the RGB modality, whereas MMPedestron is trained on a mixture of multiple modalities.
\textbf{Human-centric generalist model.} Recent studies~\cite{tang2023humanbench,ci2023unihcp,jin2024you,jin2024unifs,zeng2022not} have explored the development of a generalist model that exploits the commonalities among multiple human-centric tasks. 
For example, Tang et al.~\cite{tang2023humanbench} propose HumanBench, a human-centric benchmark comprising six downstream tasks, and train a generalist model based on this benchmark. And UniHCP~\cite{ci2023unihcp} trains a unified model for human-centric perception using a combination of 33 datasets. While HumanBench and UniHCP primarily focus on the RGB modality to obtain a model suitable for multiple tasks, our focus lies in pedestrian detection and aims to develop a model suitable for multiple modalities.

\section{MMPD Benchmark}

\begin{figure*}[t]
  \centering
    \includegraphics[width=0.8\textwidth]{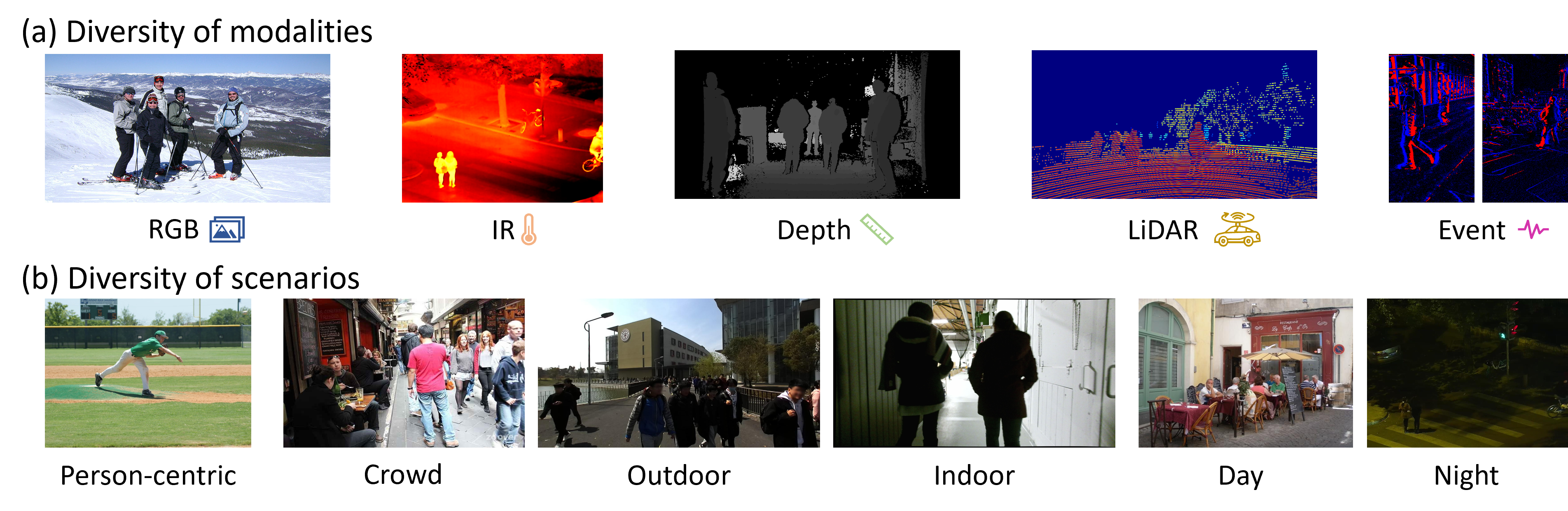}
    \vspace{-2mm}
  \caption{
  Overview of our proposed MMPD benchmark. (a) It encompasses a wide range of modalities, such as RGB, IR, Depth, LiDAR, and Event. (b) It includes diverse scenarios, including person-centric v.s. crowd, outdoor v.s. indoor, day v.s. night scenes.} 
  \label{fig:dataset}
  \vspace{-4mm}
\end{figure*}

In this paper, we introduce Multi-Modal Pedestrian Detection (MMPD) benchmark based on existing datasets and our newly collected EventPed dataset to comprehensively study the challenging task of multi-modal pedestrian detection. As depicted in Fig.~\ref{fig:dataset}, MMPD dataset offers an extensive representation of pedestrians, encompassing various modalities, including RGB, IR, depth, LiDAR, and event, and diverse scenarios, such as different types of occlusion, view-point and illumination conditions. 

\subsection{Dataset Composition}
MMPD is composed of the following datasets:
\textbf{Objects365-Persons} is derived from Objects365~\cite{shao2019objects365}, which is a large-scale RGB-based object detection dataset. We utilize only the training images related to the ``person" category.
\textbf{COCO-Persons} is a subset of COCO~\cite{lin2014microsoft,jin2020whole}, a well-known RGB-based object detection dataset, which includes images containing the "person" category.
\textbf{CrowdHuman}~\cite{shao2018crowdhuman} is a widely used RGB-based benchmark dataset for pedestrian detection in crowd scenarios.
\textbf{LLVIP}~\cite{jia2021llvip} is a visible-infrared paired dataset for low-light vision, containing $15K$ RGB-IR image pairs.
\textbf{InOutDoor}~\cite{mees2016choosing} is an RGB-Depth paired dataset for pedestrian detection. It contains $6,316$ image pairs for training and $1,028$ image pairs for evaluation.
\textbf{STCrowd}~\cite{cong2022stcrowd} is an RGB-LiDAR paired dataset for pedestrian perception in crowded scenes. It includes $5,262$ image pairs for training and $2,988$ pairs for evaluation.
\textbf{EventPed.} To address the lack of paired RGB-Event data, we propose the EventPed dataset for pedestrian detection. More details are presented in Section.~\ref{sec:dataset:eventped}.

\subsection{EventPed Dataset}
\label{sec:dataset:eventped}

EventPed dataset is a newly collected RGB-event paired dataset focusing on pedestrian detection, which can be useful for robotics, autonomous driving, and surveillance applications. It addresses the scarcity of annotated RGB-event data, facilitating future research and development.

\textbf{Data collection.}
The EventPed dataset was collected from March 2023 to July 2023.
It encompasses individuals aged between 20 and 70 years recorded in diverse outdoor scenarios such as parks and sidewalks, encompassing both day and night conditions. 
Prior to recording, informed written consent was obtained from all individuals involved.
Our portable data collection device contains a high resolution event camera~\cite{finateu20205} (IMX636), and a high-quality RGB camera (IMX586). 
The camera height varied across recordings, and event streams were captured at diverse time intervals ranging from 10 to 20 ms.

\textbf{Annotation.}
The dataset underwent manual annotation by well-trained annotators. Each individual in the dataset was exhaustively annotated with a full bounding box. In cases where individuals were partially occluded, annotators were instructed to complete the invisible parts and provide a full bounding box. Quality inspections and manual corrections are performed to ensure the annotation quality. 
We split the annotated data into a training set with $7,195$ image pairs and a test set with $2,435$ image pairs.

\subsection{Evaluation Scenarios}
A generalist multi-modal pedestrian detection model is expected to be capable of dealing with both diverse modalities and different modal combinations. So we establish two evaluation scenarios for MMPD dataset. 
\textbf{Unimodal evaluation.} We feed the model with the input in a single modality and evaluate the model on the test set of COCO, CrowdHuman, LLVIP, InOutDoor, STCrowd, and EventPed. For the datasets with multiple modalities, we report the performance with input in both modalities.
\textbf{Multi-modal evaluation.} We choose four datasets with multiple modalities, \ie LLVIP, InOutDoor, STCrowd, and EventPed, and evaluate the performance of the model with multi-modal input.
\textbf{Metrics.}
Unless otherwise specified, we use COCO AP~\cite{lin2014microsoft}(IoU=0.5:0.95, maxDets=100) as the evaluation metric on all datasets.

\section{MMPedestron Model}

\begin{figure*}[t]
  \centering
    \includegraphics[width=0.99\textwidth]{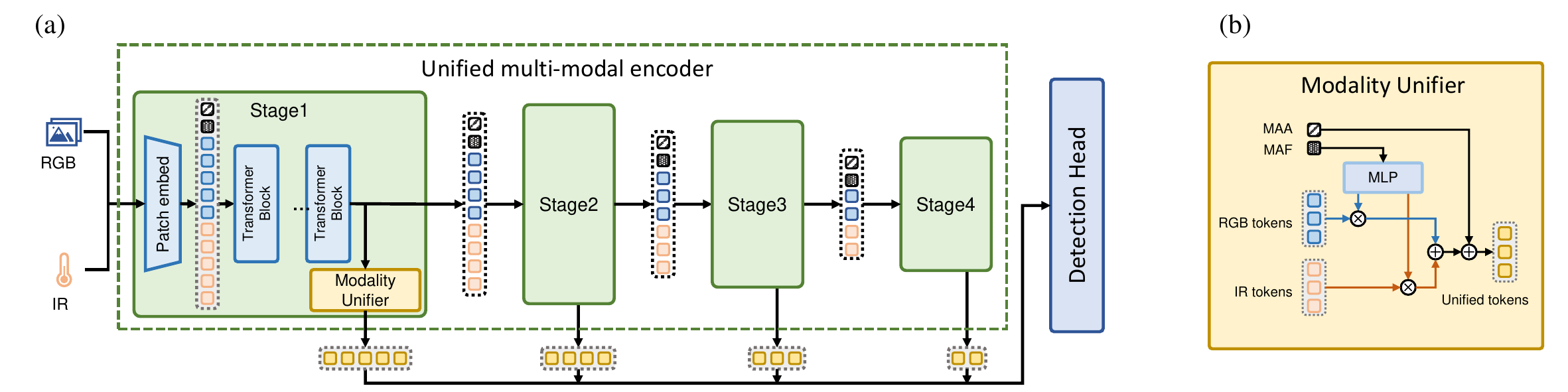}
\vspace{-2mm}
  \caption{(a) MMPedestron consists of an unified multi-modal encoder and a detection head. Each stage of the encoder contains a modality-specific patch embedding layer, several transformer blocks and a modality unifier. The resulting unified tokens from multiple stages are fed into the detection head to produce detection results. 
  (b) Modality unifier fuses multi-modal vision tokens with the guidance of MAF and incorporates the domain knowledge of MAA to the output unified tokens. For clarity, we show the case of two modalities.} 
  \label{fig:framework}
  \vspace{-5mm}
\end{figure*}

\subsection{Overview}
The overview of MMPedestron framework is illustrated in Fig.~\ref{fig:framework} (a).
MMPedestron consists of a unified multi-modal encoder and a detection head.
The unified encoder directly takes multi-modal data (\eg RGB and IR data) as the input, and generates unified vision tokens with information aggregated from multiple modalities.
These unified tokens are then fed into the detection head to obtain the final result.
Our unified tokens are compatible with various pedestrian detection heads, and in our implementation, we choose the recent Co-Dino~\cite{zong2023detrs} head for its effectiveness.

\subsection{Unified Multi-Modal Encoder}
\label{sec:mmpdestron:encoder}
In contrast to previous approaches~\cite{broedermann2022hrfuser, zhang2023cmx, zhang2023delivering} that employ separate branches for processing multi-modal data, our MMPedestron utilizes a unified transformer encoder to handle data from all modalities.
As shown in Fig.~\ref{fig:framework} (a), the unified encoder follows a multi-stage architecture with four hierarchical stages. Each stage contains a modality-specific patch embedding, a series of stacked transformer blocks, and a modality unifier.
In line with common practice~\cite{dosovitskiy2020image,liu2021swin,wang2021pyramid}, we convert the input data from each modality into a sequence of vision tokens using a modality-specific patch embedding layer within each stage. Additionally, we prepend two extra learnable tokens, \ie the modality-aware abstractor (MAA) and the modality-aware fuser (MAF), to capture the knowledge of input modality combinations.  
The multi-modal vision tokens, along with the MAA and MAF tokens, are combined to form a hybrid token sequence, which undergoes further processing by multiple transformer blocks. Our framework is compatible with various commonly used vision transformer blocks, and in this study, we employ the dual vision transformer block~\cite{yao2023dual} for its excellent performance and efficiency.
The unified multi-modal encoder offers several advantages over traditional multi-branch architectures. (1) It is more lightweight as different modalities share most of the parameters. (2) It allows the model to learn general knowledge across all modalities, enhancing its ability to generalize and adapt to diverse modalities.
(3) It enables more effective and more comprehensive message passing through the attention mechanism within each transformer block.

\subsection{Modality Unifier}
\label{sec:mmpdestron:unifier}
Given the presence of multi-modal vision tokens, conventional detection heads face difficulties in discerning the optimal utilization of these tokens. To address this issue, we propose a modality unifier module that transforms the hybrid token sequence into a unified token sequence with the same format as standard unimodal vision tokens.

Our unifier module employs two additional learnable tokens to guide the unification process.
The \textbf{Modality-Aware Fuser (MAF)} token aims to assess the importance or relevance of each modality in the multi-modal fusion process. 
\textbf{Modality-Aware Abstractor (MAA)} token aims to collect the domain knowledge related to input modalities. 

As illustrated in Fig.~\ref{fig:framework} (b), when faced with multi-modal inputs, we initially employ a  Multilayer Perceptron (MLP) to process the MAF token, obtaining confidence scores for all modalities: 
$\bm{c}=Sigmoid(MLP(\bm{x}_\text{MAF}))$,
where $\bm{x}_\text{MAF}$ denotes the feature of the MAF token, and $\bm{c}$ represents the modality confidence, reflecting the importance of each modality.
Subsequently, we fuse multi-modal vision tokens through weighted averaging, with weights determined by the predicted modality confidence $c$. Additionally, we aggregate the information contained in the MAA token by incorporating it into the unified tokens:

\begin{equation}
\label{eq:average}
\mathbf{X}_{\text{uni}} = \frac{\sum_{i=1}^{m}{(w_i * c_i * \mathbf{X}_i)}}{\sum_{i=1}^{m}{(w_i)}} + \bm{x_\text{MAA}},
\end{equation}
\begin{equation}
\label{eq:valid}
\text{where} \quad w_i= \begin{cases}
1,\quad &\text{$\mathbf{X}_i$ is valid token.} \\
0,\quad &\text{$\mathbf{X}_i$ is padded empty token.}
\end{cases} 
\end{equation}
$\mathbf{X}_{\text{uni}}$ denotes the features of the unified tokens, and $\bm{x_\text{MAA}}$ is the feature of the MAA token. $m$ is the number of modalities, $c_i$ and $X_i$ denote the confidence and token feature of the $i$th modality, respectively. $w_i$ is a factor reflecting whether the $i$th modality is valid. If a modality is missing, we pad the corresponding empty tokens. In such cases, we set $w_i$ to 0, indicating that the modality is absent. Conversely, if a modality is present, we set $w_i$ to 1, denoting its validity. More details are discussed in Section~\ref{sec:mmpedestron:train}.

The MAF token and dynamic fusion process in the unifier module allow for adaptive adjustment of the contribution of each modality based on their importance. It enables our model to adaptively allocate attention and resources to different modalities, and to leverage the complementary information provided by each modality.
The MAA token provides the domain knowledge related to input modalities. It enhances the model's ability to understand and utilize the specific characteristics of each modality.

\subsection{Multi-Modality Training}
\label{sec:mmpedestron:train}
We utilize a two-stage training scheduler for MMPedestron. 
In the RGB pretrain stage, we train our MMPedestron on the mixture of large-scale RGB datasets, including Objects365-Persons~\cite{shao2019objects365}, COCO-Persons~\cite{lin2014microsoft}, and CrowdHuman~\cite{shao2018crowdhuman}, to learn the general knowledge about human body.
In the multi-modal training stage, we train the pretrained model on the mixture of CrowdHuman~\cite{shao2018crowdhuman}, LLVIP~\cite{jia2021llvip}, InOutDoor~\cite{mees2016choosing}, STCrowd~\cite{cong2022stcrowd}, and EventPed datasets. 
The hybrid training data contains diverse modalities and modality combinations, which is essential to a generalist multi-modal model. 
We treat all modalities as 2D image inputs. For the LiDAR data, we project the 3D points into an image with sparse depth points. For the event data, we integrate the event signals with a time interval to get a 2D image. When dealing with $N$ input modalities, previous approaches typically employ $N$ independent branches. In contrast, the MAA and MAF enables MMPedestron to handle all modalities with a single shared branch, effectively reducing the parameter number by a factor of $N$.

We design a modality dropout strategy to enable our MMPedestron model to effectively handle both diverse unimodal inputs and their combinations. 
This strategy involves randomly dropping a modality from the multi-modal inputs with a probability denoted as $p$. The model is compelled to process individual modalities as well as their joint representations.
When a specific modality is missing in the input, we pad an empty image to the input and mask out all the padded tokens during the process in the encoder, including the transformer blocks and the unifier module.
For classification, we use quality focal loss~\cite{li2020generalized} and cross-entropy loss, and for regression, we employ GIOU loss~\cite{rezatofighi2019generalized} and $L_1$ loss, following established practices~\cite{zong2023detrs,zhang2022dino,zheng2022progressive}.

\section{Experiments}
We conducted a comprehensive evaluation of our MMPedestron model on multiple challenging datasets, assessing its performance through both unimodal and multi-modal fusion evaluations. Additionally, we examined the transferability of our model using cross-dataset transfer evaluation.

\subsection{Implementation Details}
MMPedestron adopts Dual ViT~\cite{yao2023dual} as the backbone, pretrained on ImageNet1K. The training process consists of two stages: \textbf{RGB Pretrain Stage}. We pretrain MMPedestron on the combined RGB-based dataset for 12 epochs using 64 NVIDIA V100 GPUs. The RGB pretrain stage requires a total of 27,648 GPU hours. \textbf{Multi-modal Training Stage}. We train MMPedestron on multi-modality datasets, with modality dropout probability $p=0.3$. The entire training process comprises 550k iterations using 32 NVIDIA V100 GPUs. The multi-modal training stage requires a total of 1,056 GPU hours.
Please refer to Supplementary for more details.

\subsection{Unimodal Evaluation}

\subsubsection{Multi-modal datasets.}
We first evaluate our MMPedestron model on various multi-modal datasets (\ie LLVIP~\cite{jia2021llvip}, STCrowd~\cite{cong2022stcrowd}, InOutDoor~\cite{mees2016choosing}, and EventPed) using the unimodal setting. 
We compare MMPedestron with a range of single-modality detectors, including two-stage detectors \ie Faster R-CNN~\cite{ren2015faster}, one-stage detectors \ie YOLOX~\cite{ge2021yolox} and query-based detectors \ie Co-Dino~\cite{zong2023detrs}. We also compare with a recent multi-modality model Meta-Transformer~\cite{zhang2023meta}. 
As illustrated in Fig.~\ref{fig:unimodal}, MMPedestron consistently outperforms the competing models across all datasets and modalities. Notably, we evaluate MMPedestron directly on the test set without further dataset-specific fine-tuning. These results demonstrate the general capability of MMPedestron in handling diverse modalities.

\subsubsection{COCO-Persons dataset.}
To validate the effectiveness of MMPedestron on traditional RGB-based detection, we compare it with state-of-the-art pedestrian detection methods on the widely-used COCO-Persons dataset~\cite{lin2014microsoft}.
We compare MMPedestron against notable models trained on COCO dataset, \ie Faster-RCNN~\cite{ren2015faster} and DINO~\cite{zhang2022dino}.
Note that these models are originally trained to handle general 80 classes (as indicated by `-80' in Table~\ref{tab:coco}). To ensure fair comparisons, we utilize MMDetection~\cite{chen2019mmdetection} to re-train these models on ``Person'' category using the default experimental setting (maked with `-Person' in Table~\ref{tab:coco}). 
For fair comparisons under similar model size (number of parameters), we report the results using the ResNet-50 as the backbone.
Additionally, we compare MMPedestron with two unified models trained on large-scale datasets: UniHCP~\cite{ci2023unihcp} and InternImage-XL~\cite{wang2023internimage}.
UniHCP is pretrained with multi-task learning on a mixture of 33 human-centric datasets, including COCO-Persons dataset. InternImage-XL model is pretrained on ImageNet22k and then finetuned on COCO dataset.
Note that we report the result of InternImage-XL with Cascade R-CNN \footnote{https://github.com/OpenGVLab/InternImage}, because InternImage-H is not publicly available.
As shown in Table~\ref{tab:coco}, our MMPedestron with direct evaluation demonstrates a significant performance margin of 3.4 AP against InternImage-XL, which is 6$\times$ larger than MMPedestron. Furthermore, fine-tuning MMPedestron on the COCO-Persons dataset further improves the performance to 71.1 AP. These remarkable results on the COCO-Persons dataset provide strong evidence of the effectiveness of MMPedestron in handling RGB data.

\begin{table*}[t]
\caption{System-level comparisons with state-of-the-art RGB-based pedestrian detection. $\dagger$ means the models with the composite techniques~\cite{liang2022cbnet}. $\uparrow$ means higher is better, while $\downarrow$ means lower is better.}
\vspace{-3mm}
\centering
\begin{minipage}[t]{0.46\textwidth}
\centering
\scalebox{0.88}{
\begin{tabular}{lcc}
  \toprule
   & \#Param. & $\text{AP}$ $\uparrow$ \\
  \midrule
  Faster RCNN-80~\cite{ren2015faster}  & 42M & 51.9 \\
  Faster RCNN-Person~\cite{ren2015faster} & 41M & 54.1 \\
  DINO-80~\cite{zhang2022dino} & 218M & 62.3   \\
  DINO-Person~\cite{zhang2022dino} & 218M & 61.8 \\
  \midrule
  UniHCP~\cite{ci2023unihcp} & 109M & 58.1  \\
  InternImage-XL~\cite{wang2023internimage} & 387M  & 64.8  \\
  \rowcolor{ourscolor}
  Ours (Direct) & 62M &  68.2 \\
  \rowcolor{ourscolor}
  Ours (Finetune) & 62M & \textbf{71.1}  \\
  \bottomrule
\end{tabular}
}
\subcaption{COCO-Persons \texttt{val} dataset}
  \label{tab:coco}
\end{minipage}
\hspace{0.04\textwidth}
\begin{minipage}[t]{0.46\textwidth}
\centering
\scalebox{0.71}{
\begin{tabular}{lcccc}
      \toprule
       & \#Param. & $\text{AP}$ $\uparrow$ & $\text{MR}^{-2}$ $\downarrow$ & $\text{JI}$ $\uparrow$   \\
      \midrule
      DETR~\cite{carion2020end} & 41M & 75.9 & 73.2 & 74.4 \\
      CrowdDet~\cite{chu2020detection} & 42M &90.7 & 41.4 & 82.3 \\
      PEDR~\cite{lin2020detr}  & 41M & 91.6& 43.7 &83.3\\
      D-DETR~\cite{zhu2020deformable} & 40M & 91.5 & 43.7 & 83.1\\
      S-RCNN~\cite{sun2021sparse} & 106M & 91.3 & 44.8 & 81.3 \\
      Iter-D-DETR~\cite{zheng2022progressive} & 207M & 94.1 & 37.7 & 87.1 \\
      \hline
      PATH~\cite{tang2023humanbench} & 320M & 90.8 & - & - \\
      UniHCP~\cite{ci2023unihcp} & 109M & 92.5  & 41.6  & 85.8 \\
      InternImage-H~\cite{wang2023internimage} & 1.09B & 
 95.4  & 37.9  &   86.6 \\
      InternImage-H$\dagger$~\cite{wang2023internimage} & 2.18B &  \textbf{97.2}  &  31.1  & \textbf{89.7}\\ 
      \rowcolor{ourscolor}
      Ours & 62M &  97.1 & \textbf{30.8} & 88.0 \\
      \bottomrule
    \end{tabular}
}
\subcaption{CrowdHuman dataset}
\label{tab:crowdhuman}
\end{minipage}
\vspace{-4pt}
\end{table*}

\subsubsection{CrowdHuman dataset.}
In order to assess the efficacy of MMPedestron in handling complex crowd scenarios, we compare our model with the state-of-the-art methods on the CrowdHuman benchmark. 
The evaluation metrics used in this benchmark include AP, $\text{MR}^{-2}$, and Jaccard index (JI), which are commonly employed in previous studies~\cite{chu2020detection}.
As presented in Table~\ref{tab:crowdhuman}, our MMPedestron model outperforms its counterparts by a substantial margin, without increasing the model complexity. Specifically, MMPedestron surpasses PATH~\cite{tang2023humanbench} (ViT-L), UniHCP~\cite{ci2023unihcp} (ViT-B), and Iter-D-Detr~\cite{zheng2022progressive} (Swin-L) by 6.3\% AP, 4.6\% AP and 3.0\% AP, respectively. Even when compared to state-of-the-art large-scale model, such as InternImage-H, which is over 30 times larger than our MMPedestron, our model achieves comparable performance. These results validate the ability of MMPedestron to handle challenging crowd scenarios.

\begin{table}[tb]
  \centering
  \caption{Multi-modality fusion evaluation. AP is reported.
  }
  \label{tab:fusion}
  \resizebox{0.75\textwidth}{!}{
    \setlength{\tabcolsep}{10pt}
    \begin{tabular}{lccccc}
      \toprule
      Method & \#Param. & LLVIP & STCrowd & InOutDoor & EventPed  \\
      \midrule
      Early-Fusion~\cite{liu2016multispectral} & 41M & 53.6 & 54.4 & 58.3 & 47.4  \\
      FPN-Fusion~\cite{tomy2022fusing} & 65M & 57.2  & 61.5 & 60.1 & 61.1  \\
      ProbEN~\cite{chen2022multimodal} & 82M & 54.8 & 60.0 & 62.4 & 60.1  \\
      HRFuser~\cite{broedermann2022hrfuser} & 101M & 53.9 & 49.0 & 58.6 & 46.0  \\
      CMX~\cite{zhang2023cmx} & 150M & 59.6  & 61.0 & 62.3 & 58.0\\ \midrule
      Ours (RGB Pretrain only) & 62M & 50.0 & 59.5 & 36.8 & 71.9 \\
      \rowcolor{ourscolor}
      Ours (RGB Pretrain + Multi-modal Training) & 62M & \textbf{72.6} & \textbf{74.9} & \textbf{65.7} & \textbf{79.0} \\
      \bottomrule
    \end{tabular}
    }
\end{table}

\subsection{Multi-Modal Evaluation}
To assess the capacity to integrate information from multiple modalities, we conducted a comparative analysis between our MMPedestron model and various modality-fusion approaches, including early-fusion~\cite{liu2016multispectral}, mid-fusion~\cite{broedermann2022hrfuser,zhang2023cmx} and late-fusion approaches~\cite{chen2022multimodal}. 
Early-Fusion~\cite{liu2016multispectral} involves concatenating multi-modal data prior to model input. 
FPN-Fusion~\cite{tomy2022fusing} fuses features from various encoder stages through simple addition. 
And CMX~\cite{zhang2023cmx} adopts a transformer-based model to incorporate the fusion of RGB with various modalities. It was originally designed for semantic segmentation, but we adapted it to perform pedestrian detection. 
HRFuser~\cite{broedermann2022hrfuser} fuses multiple sensors in a multi-resolution fashion with multi-window cross-attention blocks.
ProbEN~\cite{chen2022multimodal} trains separate models for each modality individually and aggregates the predicted bounding boxes of all models with post-processing.
As depicted in Table~\ref{tab:fusion}, when compared to models separately trained on specific datasets, our MMPedestron model consistently outperforms all fusion methods on diverse datasets without requiring dataset-specific fine-tuning. These results validate the efficacy of MMPedestron in integrating multi-modal information and its ability to generalize across various modality combinations. Moreover, we report the results of our MMPedestron in the RGB pretrain stage, which only uses the RGB modality input. Although RGB pretrain could help learn the general knowledge about human body, multi-modal training significantly improves the detection performance and helps to learn extensive representation of pedestrians, in Table~\ref{tab:fusion}. This result also demonstrates the necessity of research on multi-modal pedestrian detection.

\subsection{Cross-Dataset Transfer Evaluation}
Cross-dataset transfer evaluation aims to measure the model's capability to adapt to new scenarios. We finetune our MMPedestron on the PEDRo~\cite{bbpprsPedro2023} dataset and compare it with state-of-the-art methods specifically trained on that dataset. 
It is important to note that the PEDRo dataset utilizes a distinct representation for event data, rendering our pre-trained MMPedestron model incompatible for direct evaluation. 
The results in Table~\ref{tab:pedro}, demonstrate that even with only $10\%$ of the training data, our MMPedestron model achieves state-of-the-art performance on both Val and Test sets. Furthermore, by fine-tuning MMPedestron on the complete training set, we obtain even higher performance, achieving a new state-of-the-art result of $81.5$ AP on the Val set and $73.3$ AP on the Test set.
The quick adaptation to PEDRo dataset showcases the exceptional generalization capacity of our MMPedestron model, which is a result of its multi-modal training on MMPD dataset. 

We perform multi-modality fine-tuning experiments on FLIR dataset~\cite{FLIR_dataset}. Following~\cite{zhang2023cmx, qingyun2022cross}, we conduct experiments on the ``aligned'' FLIR dataset~\cite{zhang2020multispectral}, which consists of 4,192 training pairs and 1,013 testing pairs covering three object categories: `person', `bicycle', and `car'. We extend MMPedestron to support all three categories in FLIR, which further validates its generalization and transfer ability to novel tasks and domains. As shown in the Table~\ref{tab:FLIR}, MMPedestron achieves state-of-the-art performance on the FLIR test set.

\begin{table*}[t]
\caption{Cross-dataset transfer evaluation.}
\vspace{-2mm}
\centering
\begin{minipage}[t]{0.52\textwidth}
\centering
\scalebox{0.68}{
\setlength{\tabcolsep}{6pt}
\begin{tabular}{lc|cc}
  \toprule
   Method & \#Param. & Val AP & Test AP \\
  \midrule
  YOLOv8~\cite{Jocher_YOLO_by_Ultralytics_2023}  & 68M & - & 58.6 \\
  Faster R-CNN~\cite{ren2015faster}  & 41M & 66.8 & 58.0 \\
  YOLOX~\cite{ge2021yolox} & 99M & 73.9 & 68.8 \\
  Co-Dino~\cite{zong2023detrs} & 64M & 73.4 & 65.4 \\ 
  Meta Transformer~\cite{zhang2023meta} & 155M & 67.5 & 61.1 \\ 
  \midrule 
  \rowcolor{ourscolor}
  Ours (10\% train data) & 62M & 79.3	& 72.7 \\
  \rowcolor{ourscolor}
  Ours & 62M & \textbf{81.5} & \textbf{73.3} \\
  \bottomrule
\end{tabular}
}
\subcaption{Pedestrian detection on PEDRo dataset (Event).}
  \label{tab:pedro}
\end{minipage}
\hspace{0.01\textwidth}
\begin{minipage}[t]{0.4\textwidth}
\centering
\scalebox{0.82}{
\setlength{\tabcolsep}{10pt}
  \begin{tabular}{lc|c}
      \toprule
      Method & \#Param. & $\text{AP50}$  \\
      \midrule
      GAFF~\cite{zhang2021guided} & - & 72.9 \\
      CFT~\cite{qingyun2022cross} & 208M & 77.7 \\
      CMX~\cite{zhang2023cmx} & 150M & 82.2 \\
     \rowcolor{ourscolor}
      Ours & 62M & \textbf{86.4}  \\
      \bottomrule
    \end{tabular}
}
\subcaption{Multi-category object detection on FLIR test dataset (RGB + IR).}
\label{tab:FLIR}
\end{minipage}
\vspace{-4pt}
\end{table*}

\subsection{Ablation Study}

\begin{table}[tb]
  \centering
  \caption{Ablation study on the LLVIP dataset. 
  }
  \vspace{-2mm}
  \label{tab:ablation}
  \resizebox{0.65\textwidth}{!}{
    \setlength{\tabcolsep}{10pt}
    \begin{tabular}{cc|ccc}
      \toprule
      MAF & MAA & Fusion & RGB & IR  \\
      \midrule
     \xmark & \xmark & 61.7 & 47.5 & 61.2 \\
     \cmark & \xmark & 68.2 (+6.5) & 49.8 (+2.3) & 68.2 (+7.0) \\
     \cmark & \cmark & \textbf{68.9}  (+0.7) & \textbf{52.9}  (+3.1) & \textbf{68.4}  (+0.2)\\
      \bottomrule
    \end{tabular}
    }
    \vspace{-3mm}
\end{table}

We conduct ablation study on the LLVIP dataset to comprehensively assess the effect of each component in our proposed model in Table~\ref{tab:ablation}.
The baseline (row \#1) utilizes the Dual-ViT backbone with the Co-DINO head. By comparing the performance of various configurations, we observed significant improvements when incorporating Modality-Aware Fusion (MAF). Particularly, the Fusion results showed a notable increase of 6.5 AP, showcasing the effectiveness of MAF in enhancing multi-modality fusion. Additionally, the Modality Attention Abstractor (MAA) component was found to enhance the unification of different modalities. This led to a significant improvement in the lower-performance modality (RGB), with an increase of 3.1 AP. Consequently, the fusion results were further elevated. These findings highlight the importance of both MAF and MAA in our model, as they contribute to improved performance by effectively addressing the challenges associated with multi-modal learning.

\begin{figure*}[t]
  \centering
    \includegraphics[width=0.99\textwidth]{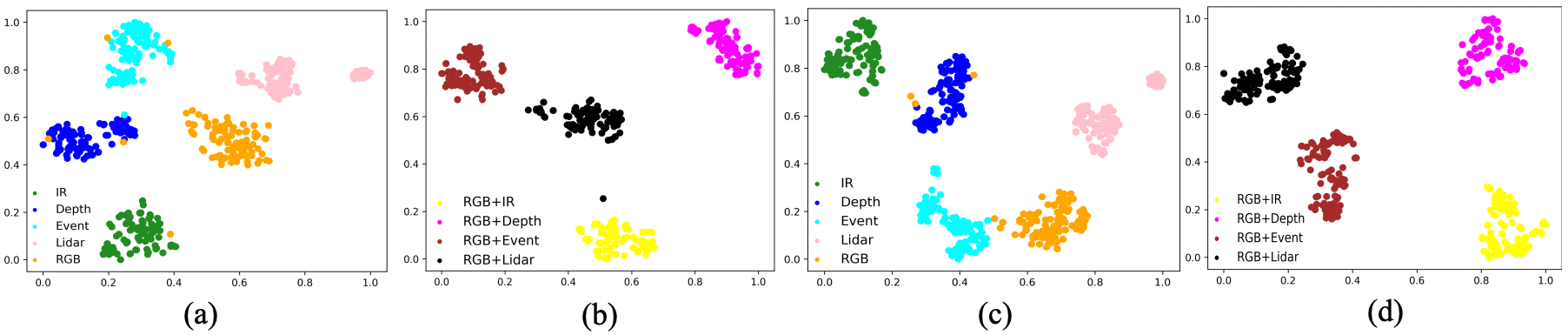}
  \vspace{-2mm}
  \caption{Visualization of MAF (a, b) and MAA (c, d) tokens. (a,c) are for unimodal inputs, and (b,d) are for multi-modal inputs.} 
  \vspace{-5mm}
  \label{fig:MA_analysis}
\end{figure*}

\subsection{Analysis of MAF and MAA}
To gain insights into the properties of our Modality-Aware Fusion (MAF) and Modality-Aware Abstractor (MAA) tokens, we conducted an analysis by visualizing distributions of token features for different input modalities with t-SNE~\cite{cieslak2020t}.
As shown in Fig.~\ref{fig:MA_analysis} (a) and (c), when processing single-modality inputs, both MAF and MAA samples show clustering patterns corresponding to the respective input modality. This indicates that our MAF and MAA are ``modality-aware'', and are able to adaptively adjust token features according to the input modality. 
Similarly, Fig.~\ref{fig:MA_analysis} (b) and (d) demonstrate distinguishable clustered patterns for different modality combinations. 
This adaptability allows our MMPedestron model to dynamically select appropriate strategies for multi-modal feature fusion according to the specific modality combination.
The modality-aware token features ensure the generalization ability of MMPedestron to diverse modalities and modality combinations.

\subsection{Runtime Analysis}
We conduct the runtime analysis of MMPedstron with a batchsize of 1 on one RTX-4090 GPU in a single thread. MMPedestron achieves near real-time performance (24fps) on the FLIR~\cite{FLIR_dataset} dataset. Model compilation, pruning, and quantization can further enhance speed, but they are beyond this paper's scope.

\section{Conclusion}
In this paper, we have presented a pioneering approach to multi-modal pedestrian detection. We introduced the MMPD benchmark, which is the first large-scale benchmark specifically designed for developing and evaluating multi-modal pedestrian detection models. Building upon this benchmark, we proposed MMPedestron, which effectively handles diverse modalities and scenarios. Through comprehensive experiments, we have demonstrated the effectiveness of our MMPedestron model. We hope our work could serve as a valuable foundation for the development of future multi-modal generalist pedestrian detection models.

\noindent\textbf{Acknowledgement.}
This paper is partially supported by the National Key R\&D Program of China No.2022ZD0161000 and the General Research Fund of Hong Kong No.17200622 and 17209324.

%
%
\bibliographystyle{splncs04}
\bibliography{egbib}

\clearpage
\appendix

\setcounter{table}{0}
\renewcommand{\thetable}{A\arabic{table}}
\setcounter{figure}{0}
\renewcommand{\thefigure}{A\arabic{figure}}

\section{EventPed Dataset}

In this section, we provide more details and analysis for our newly introduced EventPed Dataset. We introduce more details about data processing in Section~\ref{sec:supp_data_processing} and provides analysis and statistics of the dataset in Section~\ref{sec:supp_statistics}.

\subsection{Data processing.}
\label{sec:supp_data_processing} 
Time synchronization is achieved using a physical connection between the cameras. Given the small distance between the cameras, spatial registration can be approximated using a homography matrix. Initially, we performed camera calibration to estimate the intrinsic and extrinsic camera parameters. Subsequently, the homography matrix was calculated to project the coordinates of event images onto the corresponding RGB images. This process ensured the alignment of RGB-Event image pairs. Finally, we cropped the central region of the registered RGB-Event image pairs to ensure the same field of vision and image size, resulting in a resolution of $960\times512$ pixels for all samples.

\subsection{Analysis and Statistics.}
\label{sec:supp_statistics} 
\begin{figure}
    \centering
    \includegraphics[width=0.7\textwidth]{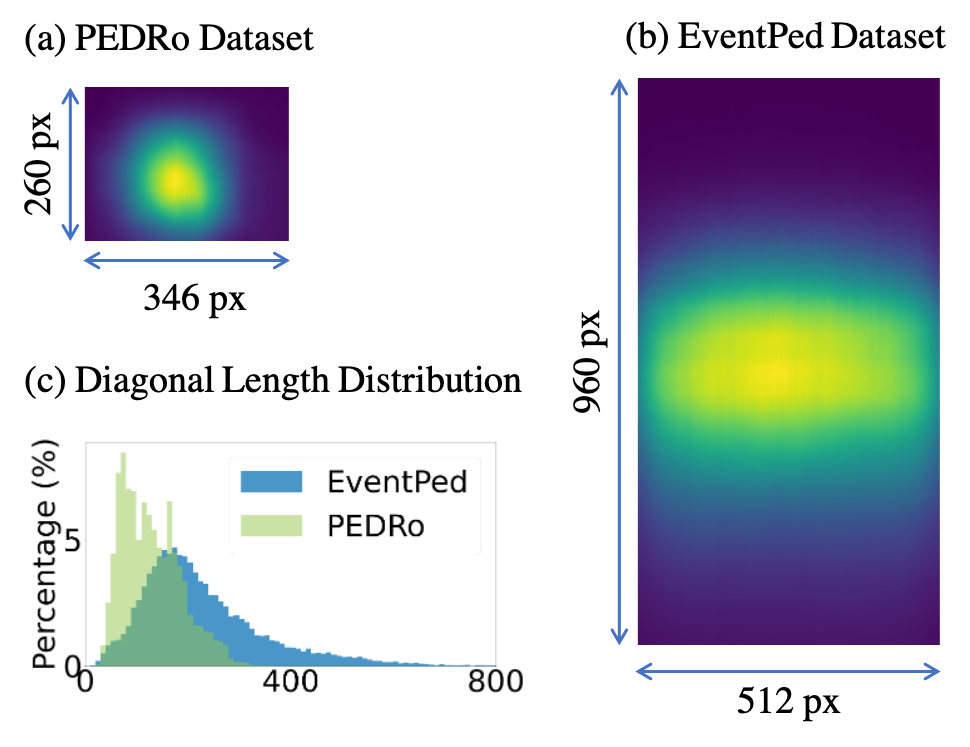}
    \caption{(a,b) The bounding box distribution of PEDRo dataset (a) and our EventPed dataset (b). The bounding box distribution of EventPed is more uniform horizontally and more concentrated vertically.
    (c) The distribution of the diagonal length in pixel of the bounding boxes in PEDRo (green) and our EventPed (blue). EventPed contains more bounding boxes in large size.}
    \label{fig:statistics_bbox}
\end{figure}

We conducted an analysis comparing the EventPed dataset with the existing event-based dataset, PEDRo. In Fig.~\ref{fig:statistics_bbox}(a,b), we present heatmaps visualizing the distribution of bounding boxes. The heatmaps illustrate the probability of bounding boxes covering each pixel. The EventPed dataset shows a more uniform distribution in the horizontal direction and a more concentrated distribution in the vertical direction compared to PEDRo.
Furthermore, we examined the distribution of bounding box sizes, as shown in Fig.~\ref{fig:statistics_bbox}(c). Our EventPed dataset exhibits a higher number of bounding boxes in larger sizes. This difference can be attributed to the disparity in image resolutions between the two datasets. The PEDRo dataset has a relatively low image resolution of $260 \times 346$, while our EventPed dataset boasts a higher resolution of $960 \times 512$.

\section{Dataset Distribution}
\begin{figure}[ht]
  \centering
  \includegraphics[width=0.9\textwidth]{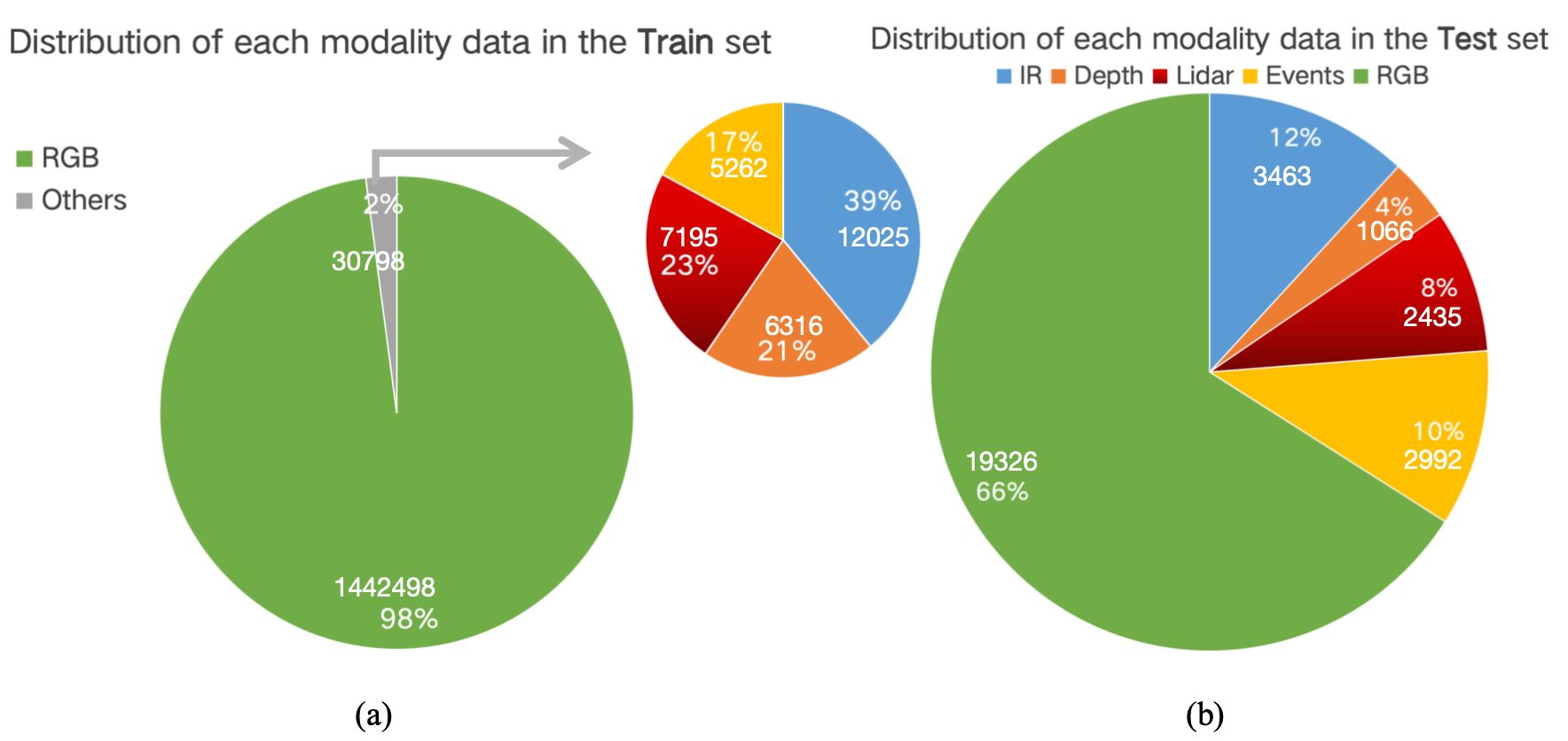}
  \caption{Distribution of different modalities: (a) Train (b) Test.}
  \label{fig:distrbution}
\end{figure}

We illustrate distribution of different modalities of MMPD dataset in Figure~\ref{fig:distrbution}. From the figure, we see the vast majority of the MMPD dataset is in the RGB modality (for RGB pretrain), with other modalities accounting for a smaller proportion, only about 2\% for training. For testing, the proportions of the different modalities are more balanced. Aside from the RGB modality, the proportions of the other modalities are relatively balanced.

\section{Experiments with the Faster R-CNN head}

To validate the applicability of our method on a wider range of detection approaches, we also implement our MMPedestron on the common Faster R-CNN~\cite{ren2015faster} head. 
Specifically, in order to improve the generalization ability across diverse scenarios, we use a wider range of box ratio when generating candidate boxes. Specifically, we set the bounding box ratio to $0.5, 1.0, 1.5, 2.0, 2.5, 3.0$. In addition, we use more proposal boxes (up to 2000 per image) to handle crowded scenarios. Please refer to Sec.5.1 in the main paper for other implementation details. 

The results are reported in Table \ref{tab:fusion_full}. Compared to models separately trained on specific datasets, our MMPedestron model consistently outperforms all fusion methods on diverse datasets without requiring dataset-specific fine-tuning. More importantly, our MMPedestron model with the Faster R-CNN detection head has achieved significantly higher performance than other methods, without introducing more parameter counts (41M only).

\begin{table}[tb]
  \centering
  \caption{Multi-modality fusion evaluation. AP is reported.
  }
  \label{tab:fusion_full}
  \resizebox{0.99 \textwidth}{!}{
    \setlength{\tabcolsep}{10pt}
    \begin{tabular}{lccccc}
      \toprule
      Method & \#Param. & LLVIP & STCrowd & InOutDoor & EventPed  \\
      \midrule
      Early-Fusion~\cite{liu2016multispectral} & 41M & 53.6 & 54.4 & 58.3 & 47.4  \\
      FPN-Fusion~\cite{tomy2022fusing} & 65M & 57.2  & 61.5 & 60.1 & 61.1  \\
      ProbEN~\cite{chen2022multimodal} & 82M & 54.8 & 60.0 & 62.4 & 60.1  \\
      HRFuser~\cite{broedermann2022hrfuser} & 101M & 53.9 & 49.0 & 58.6 & 46.0  \\
      CMX~\cite{zhang2023cmx} & 150M & 59.6  & 61.0 & 62.3 & 58.0\\ \midrule
      \rowcolor{ourscolor}
      Ours (Faster R-CNN) & 41M & 66.8 & 66.4 & 66.1 & 75.3 \\
      \rowcolor{ourscolor}
      Ours (Co-Dino) & 62M & 72.6 & 74.9 & 65.7 & 79.0 \\
      \bottomrule
    \end{tabular}
    }
\end{table}

\section{Implementation Details}

MMPedestron adopts Dual ViT~\cite{yao2023dual} as the backbone (if not explicitly indicated), pretrained on ImageNet1K. The training process consists of two stages: RGB pretrain stage and multi-modal training stage. For both stages, we employ the AdamW optimizer with parameters $\beta_1$ = 0.9, $\beta_2$ = 0.999, and weight decay set to 1e-4. The learning rate is warmed up linearly for the first 500 iterations to 1e-3 and then decayed by a factor of 0.1 at epoch 8 and 10. The training input image resolution is randomly selected between $480\times1333$ and $800\times1333$.

\subsection{RGB Pretrain Stage}
We pretrain MMPedestron on the combined RGB-based dataset for 12 epochs.
The drop path rate is set to 0.15 for the MMPedestron encoder. Training is conducted using 64 NVIDIA V100 GPUs with a total batch size of 64. The RGB pretrain stage requires a total of 27,648 GPU hours.

\subsection{Multi-modal Training Stage}
During training, we use a drop path rate of 0.3 and apply a layer-wise learning rate decay of 0.1 for the MMPedestron encoder. To enhance robustness to missing modalities, we set the modality dropout probability as $p=0.3$.
The entire training process comprises 550k iterations,  approximately equivalent to 12 epochs.
Training is conducted using 32 NVIDIA V100 GPUs, with a total batch size of 32. The multi-modal training stage requires a total of 1,056 GPU hours.

\section{Details about Baselines}

\subsection{Unimodal Baselines}
We compare our proposed MMPedestron model with a range of single-modality detectors, including two-stage detectors \eg Faster R-CNN~\cite{ren2015faster}, one-stage detectors \eg YOLOX~\cite{ge2021yolox} and query-based detectors \eg Co-Dino~\cite{zong2023detrs}  on our proposed MMPD benchmark.

\subsubsection{Experiments on multi-modal datasets}

\textbf{Faster R-CNN}\cite{ren2015faster} is a two-stage, anchor-based detector. For comparisons, we reported the results of the model with the ResNet50~\cite{he2016deep} backbone. The model is trained with MMDetection~\cite{chen2019mmdetection} using the default 1x training setting (12 epochs)\footnote{https://github.com/open-mmlab/mmdetection/}.

\textbf{YOLOX}\cite{ge2021yolox} is a single-stage, anchor-based detector. For comparisons, we reported the results of YOLOX-X, which is the largest one in the YOLOX series. All hyper-parameters are set to the default values of YOLOX~\cite{ge2021yolox} in MMDetection~\cite{chen2019mmdetection}. For example, the total training epochs is 300. 

\textbf{Co-Dino}\cite{zong2023detrs} is a collaborative hybrid, query-based detector. For comparisons, we reported the results of the model with the ResNet50~\cite{he2016deep} backbone. All hyper-parameters are set to the default values of Co-Dino~\cite{zong2023detrs} in MMDetection~\cite{chen2019mmdetection}. 

\textbf{Meta-Transformer}\cite{zhang2023meta} is a unified multimodal encoder which can handle various data formats, such as natural language, images, audio, point clouds, etc. It also supports various downstream tasks, such as classification, segmentation, object detection, etc. For comparisons, we choose the Meta-Transformer-B16 base-scale model (with LAION-2B\cite{radford2021learning} pretraining) as the backbone. We follow ~\cite{zhang2023meta} to adopt ViT-adapter techniques for training and use Cascade RCNN head for the detection task. We directly use the official codes\footnote{https://github.com/invictus717/MetaTransformer/tree/master/Image/detection} without changing any hyper-parameters.

\subsubsection{Experiments on COCO-Persons dataset}

We compare MMPedestron against notable models trained on COCO dataset, \ie Faster-RCNN~\cite{ren2015faster} and DINO~\cite{zhang2022dino}.
Note that these models are originally trained to handle general 80 classes (as indicated by `-80'). To ensure fair comparisons, we utilize MMDetection~\cite{chen2019mmdetection} to re-train these models on ``Person'' category using the default experimental setting. Note that MMDetection re-implementation can be a little bit better than the original implementation. 

\textbf{Faster R-CNN}~\cite{ren2015faster} is a two-stage, anchor-based detector. For fair comparisons under similar model size (number of parameters), we report the results using the ResNet-50 as the backbone. All hyper-parameters are set to the default values in MMDetection~\cite{chen2019mmdetection}. 

\textbf{DINO}~\cite{zhang2022dino} is an query based with improved deNoising anchor boxes end-to-end object detection model. 
For fair comparisons under similar model size (number of parameters), we report the results using the ResNet-50 as the backbone. The models are trained for 50 epochs. Other hyper-parameters are set to the default values in MMDetection~\cite{chen2019mmdetection}. 

\textbf{UniHCP}~\cite{ci2023unihcp} is a unified model for human-centric perceptions, which uses the standard ViT-L~\cite{dosovitskiy2020image} as the encoder network. The model is pre-trained with the MAE~\cite{he2022masked} techniques, and then trained with multi-task learning on a mixture of 33 human-centric datasets, including COCO-Persons dataset. The results are from the authors~\cite{ci2023unihcp}.

\textbf{InternImage}~\cite{wang2023internimage} is a multimodal multitask general large-scale foundation models with deformable convolutions. In the experiments, we compare with the result of InternImage-XL with Cascade R-CNN \footnote{https://github.com/OpenGVLab/InternImage}, because InternImage-H model is not publicly available. InternImage-XL is pretrained on ImageNet22k and then finetuned on COCO dataset.

\subsubsection{Experiments on CrowdHuman dataset}

\textbf{DETR}\cite{carion2020end} is an end-to-end detection model which views object detection as a direct set prediction problem. For comparisons, we reported the results of the model with the ResNet50~\cite{he2016deep} backbone. The results are from the paper~\cite{zheng2022progressive}.

\textbf{CrowdDet}\cite{chu2020detection} is an object detection model based on Faster-R-CNN, and the refinement module enables it to effectively handle the difficulty of detecting highly overlapping objects. For comparisons, we reported the results of the model with the ResNet50~\cite{he2016deep} backbone. The results are from the paper~\cite{zheng2022progressive}. 

\textbf{PEDR}\cite{lin2020detr} is a DETR\cite{carion2020end} based detector, and the dense queries and corrected attention field (DQRF) decoder significantly improves the performance of the model in crowded scenarios. For comparisons, we reported the results of the model with the ResNet50~\cite{he2016deep} backbone. The results are from the paper~\cite{zheng2022progressive}. 

\textbf{Deformable DETR(D-DETR)}\cite{zhu2020deformable} is an query based detector. The additional deformable attention module can be naturally extended to aggregating multi-scale features, without the help of FPN\cite{lin2017feature}. For comparisons, we reported the results of the model with the ResNet50~\cite{he2016deep} backbone. The results are from the paper~\cite{zheng2022progressive}. 

\textbf{Sparse-RCNN(S-RCNN)}\cite{sun2021sparse} is an end-to-end Object Detection with learnable proposals. For comparisons, we reported the results of the model with the ResNet50~\cite{he2016deep} backbone. The results are from the paper~\cite{zheng2022progressive}. 

\textbf{Iter-Deformable-DETR(Iter-D-DETR)}\cite{zheng2022progressive} is a progressive Deformable-DETR based method equipped with a prediction selector, relation information extractor, query updater, and label assignment to improve the performance of query-based object detectors in handling crowded scenes. In our comparison, we chose the model with the Swin-L~\cite{liu2021swin} backbone. The results are from the original paper~\cite{zheng2022progressive}. 

\textbf{PATH}~\cite{tang2023humanbench} is a general human-centric perception model based on the standard ViT-L~\cite{dosovitskiy2020image}. Projector-Assisted Hierarchical pretraining method (PATH) can be used to learn diverse knowledge at different granularity levels. The model is pre-trained on a collection of 11,019,187 pretraining images from 37 human-centric datasets, and then fine-tuned on the COCO-Persons dataset. The results are from the original paper~\cite{tang2023humanbench}.

\textbf{InternImage}~\cite{wang2023internimage} is a multimodal multitask general large-scale foundation models with deformable convolutions. For experiments on the CrowdHuman dataset, we reported the results of InternImage-H with (denoted as $^\dagger$) and without the composite techniques. The composite technique improves the model performance but at the cost of doubled model parameters. The results are obtained from the authors~\cite{wang2023internimage}.

\subsection{Multi-modal Fusion Baselines}

We compare our proposed MMPedestron model with a range of multi-modality fusion approaches, including early-fusion~\cite{liu2016multispectral}, mid-fusion~\cite{broedermann2022hrfuser,zhang2023cmx} and late-fusion approaches~\cite{chen2022multimodal}. 

\textbf{Early-Fusion}~\cite{liu2016multispectral} involves concatenating multi-modal data prior to model input. The model is based on Faster RCNN with the ResNet-50 backbone. The model is trained with MMDetection~\cite{chen2019mmdetection} using the default 1x training setting (12 epochs). 

\textbf{FPN-Fusion}~\cite{tomy2022fusing} fuses features from various encoder stages through simple addition. The model is based on Faster RCNN with the ResNet-50 backbone. The model is trained with MMDetection~\cite{chen2019mmdetection} using the default 1x training setting (12 epochs). 

\textbf{ProbEN}~\cite{chen2022multimodal} trains separate models for each modality individually and aggregate the predicted bounding boxes of all models with post-processing. The model is based on Faster RCNN with the ResNet-50 backbone. The model is trained with MMDetection~\cite{chen2019mmdetection} using the default 1x training setting (12 epochs). 

\textbf{CMX}\cite{zhang2023cmx} adopts a transformer-based model to incorporate the fusion of RGB with various modalities. It was originally designed for semantic segmentation, but we adapted it to perform pedestrian detection by introducing the Faster R-CNN based detection head. For comparisons, we choose the Swin Transformer (ImageNet\cite{deng2009imagenet} pre-trained)\cite{liu2021swin} small version backbone. The model is trained with MMDetection~\cite{chen2019mmdetection} using the default 1x training setting (12 epochs). 

\textbf{HRFuser}\cite{broedermann2022hrfuser} is a multi-resolution sensor modality fusion model. For comparisons, we choose the base-scale HRFuser model with the Cascade-RCNN detection head. We directly follow the official codes\footnote{https://github.com/timbroed/HRFuser} for model training (the default 12-epoch training schedule), without modifying the hyper-parameters.

\section{Qualitative Results}

\begin{figure*}[t]
  \centering
    \includegraphics[width=0.95\textwidth,height=0.95\textheight]{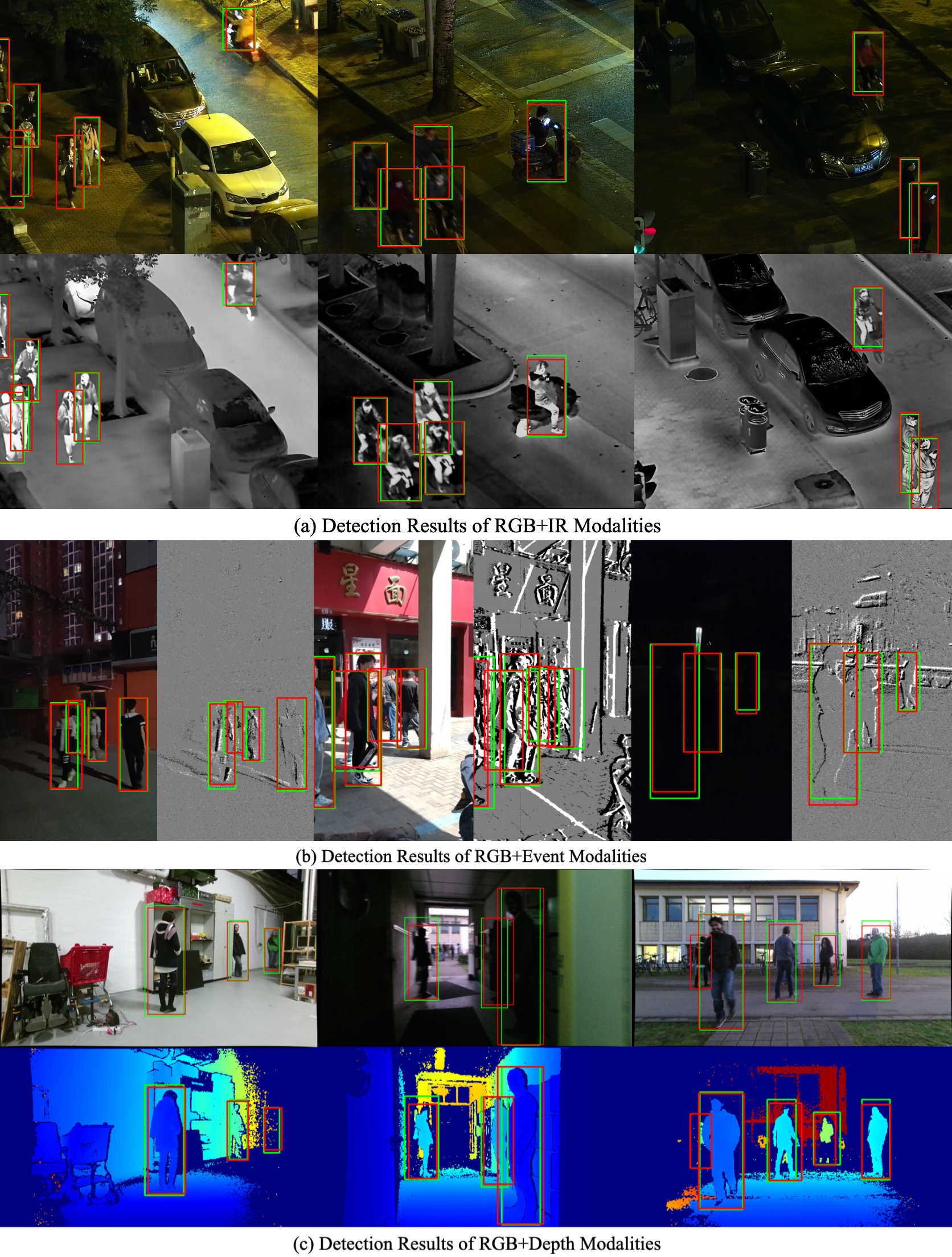}
  \caption{Visualization of RGB+IR modalities detection results (a), RGB+Event modalities detection results (b), RGB+Depth modalities detection results (c).} 
  \label{fig:fusion_all_p1}
\end{figure*}

\begin{figure*}[t]
  \centering
    \includegraphics[width=0.95\textwidth,height=0.95\textheight]{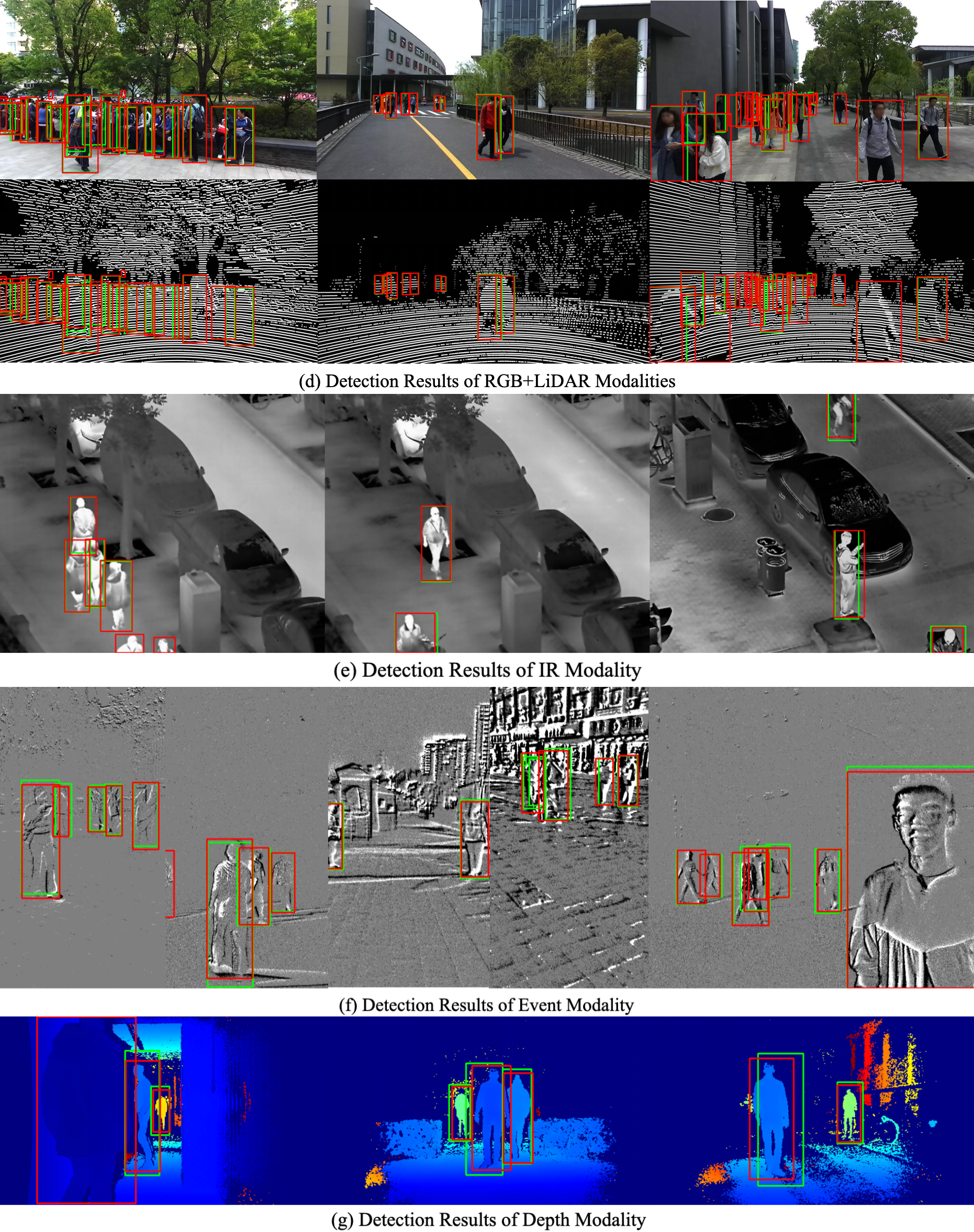}
  \caption{Visualization of RGB+LiDAR modalities detection results (d), IR modality detection results (e), Event modality detection results (f), Depth modality detection results (g).} 
  \label{fig:fusion_p2_new}
\end{figure*}

\begin{figure*}
  \centering
    \includegraphics[width=0.95\textwidth,height=0.95\textheight]{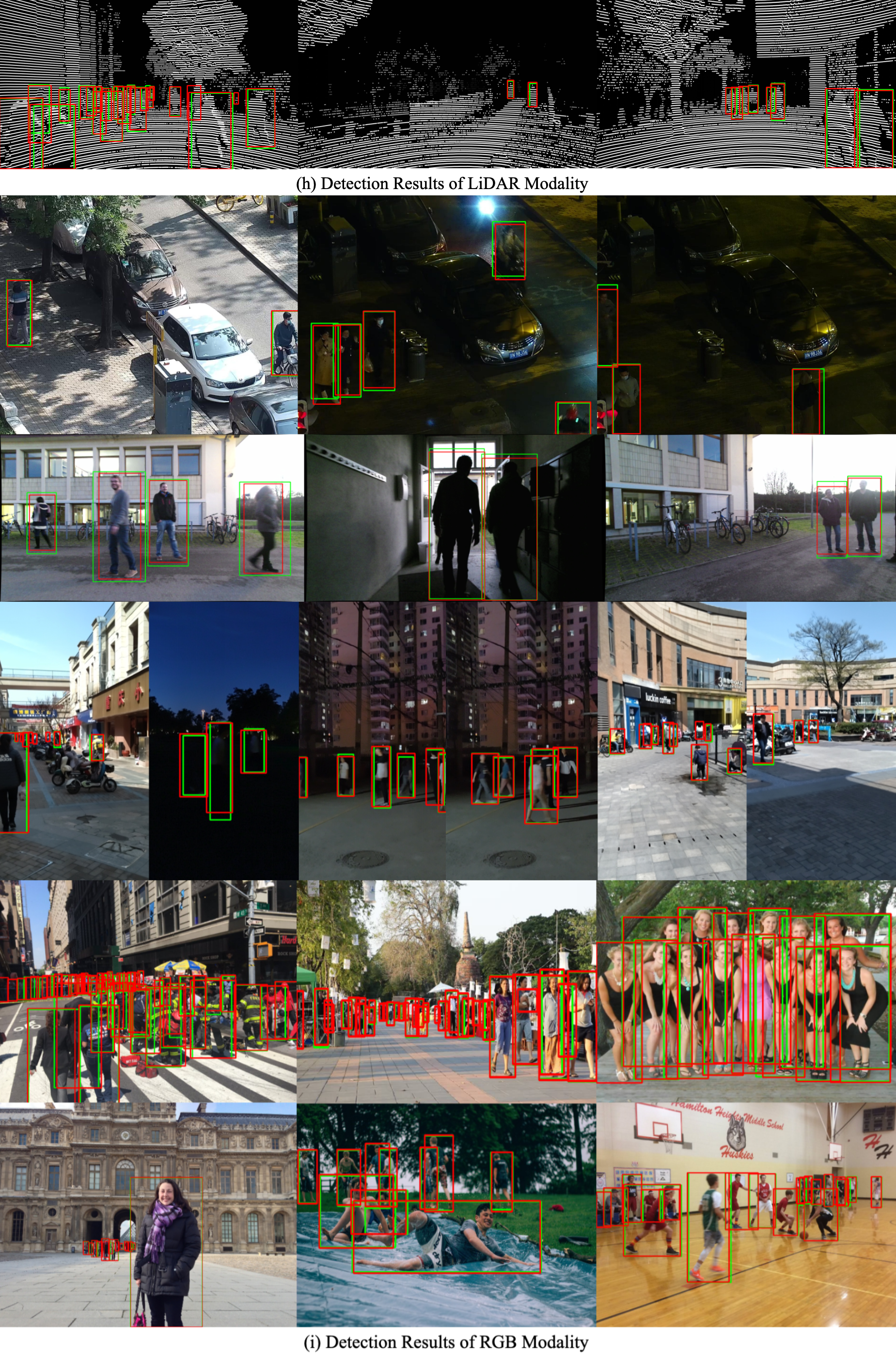}
  \caption{Visualization of LiDAR modality detection results (h), RGB modality detection results (i).} 
  \label{fig:rgb_all_p2_new}
\end{figure*}

In this section, we show some qualitative results of our proposed MMPedestron in Fig.~\ref{fig:fusion_all_p1}, Fig.~\ref{fig:fusion_p2_new} and Fig.~\ref{fig:rgb_all_p2_new}. The qualitative evaluation includes multi-modal evaluation (a-d) and unimodal evaluation (e-i). For multi-modal evaluation, we visualize the results of (a) RGB+IR, (b) RGB+Event, (c) RGB+Depth, and (d) RGB+LiDAR. And for single-modal evaluation, we visualize the results of (e) IR, (f) Event, (g) Depth, (h) LiDAR, and (i) RGB. In the images, the red boxes represent the model prediction results, and the green boxes represent the ground-truth bounding box results. From the visualization results, we can observe our MMPedestron model can handle a variety of modalities and their dynamic combinations. In addition, it shows strong generalization ability across diverse scenarios, such as different scales, types of occlusion, view-point and illumination conditions. This suggests that our MMPedestron model has good potential for a wide range of application scenarios.

\section{Limitations}

While our work focuses on 2D modalities, such as RGB and IR, other modalities such as 3D point cloud and sequence inputs like videos also hold significant potential for pedestrian detection. We encourage future research to expand upon our work by incorporating more modalities for multi-modal pedestrian detection.

\end{document}